%% file: example.tex
\documentclass[letterpaper]{article}

\usepackage{natbib,alifeconf}  
\usepackage{url,hyperref,cleveref}
\usepackage{booktabs}
\newcommand{\chatgpt}{GPT-3.5 turbo }
\newcommand{\llama}{Llama 2 }
\newcommand{\la}{LA2 }

\usepackage{comment}

\title{ALIFE2024 template}

\title{Collective Innovation in Groups of Large Language Models}

    \author{
    Eleni Nisioti$^{1}$,
    Sebastian Risi $^{1}$, 
    Ida Momennejad $^{2}$,
    Pierre-Yves Oudeyer $^{3}$ \and
    Clément Moulin-Frier $^{3}$
    \mbox{}\\
    $^1$  IT University, Denmark \\
    $^2$ Microsoft Research, United States \\
    $^3$ Inria Center of the University of Bordeaux, France \\
    enis@itu.dk
} 

\begin{document}

\maketitle

\begin{abstract}
    Human culture relies on collective innovation: our ability to continuously explore how existing elements in our environment can be combined to create new ones.
    Language is hypothesized to play a key role in human culture, driving individual cognitive capacities and shaping communication.
    Yet the majority of models of collective innovation assign no cognitive capacities or language abilities to agents.
    Here, we contribute a computational study of collective innovation where agents are Large Language Models (LLMs) that play Little Alchemy 2, a creative video game originally developed for humans that, as we argue, captures useful aspects of innovation landscapes not present in previous test-beds.
    We, first, study an LLM in isolation and discover that it exhibits both useful skills and crucial limitations.
    We, then, study groups of LLMs that share information related to their behaviour and focus on the effect of social connectivity on collective performance.
    In agreement with previous human and computational studies, we observe that groups with dynamic connectivity out-compete fully-connected groups.
    Our work reveals opportunities and challenges for future studies of collective innovation that are becoming increasingly relevant as Generative Artificial Intelligence algorithms and humans innovate alongside each other.
\end{abstract}

\input{sections/introduction}

\input{sections/related}

\input{sections/testbed}

\input{sections/background}

\input{sections/collective_LLM}

\input{sections/results}

\input{sections/discussion}

\section*{Acknowlegements}
{This research was partially funded by the French National Research Agency (\url{https://anr.fr/}, project ECOCURL, Grant ANR-20-CE23-0006) and benefitted from access to the Jean Zay (Idris) supercomputer associated with the Genci grant A0151011996.}

\footnotesize
\bibliographystyle{apalike}
\bibliography{example} 

\end{document}

%% file: sections/introduction.tex
\section{Introduction}

Human culture evolves through the accumulation of artefacts, semantic repertoires, and behaviours that become more complex over long time-scales~\citep{creanza_cultural_2017,solee_evolutionary_2013,whiten_emergence_2021}.
While popular narratives often emphasize the contributions of lonely geniuses~\citep{montuori_deconstructing_1995}, a different story emerges if we consider historical data on cultural artefacts~\citep{solee_evolutionary_2013,eldredge_paleontology_2011,pereira_shaping_2023}, theoretical analysis ~\citep{creanza_cultural_2017,solee_evolutionary_2013}, human behavioural experiments ~\citep{derexPartialConnectivityIncreases2016,masonPropagationInnovationsNetworked2008,brackbill_impact_2020} and computational  studies ~\citep{lazerNetworkStructureExploration2007,cantor2021SocialNetworkArchitecture,brackbillNetworkStructureCollective2017,fangBalancingExplorationExploitation2010}. 
This body of work suggests that human cultural evolution is an inherently collective process, akin to biological evolution~\citep{creanza_cultural_2017,solee_evolutionary_2013}:
innovations arise in a collective as individuals modify and recombine existing ones in their environment.
In this work, we contribute novel computational evidence to support this hypothesis. 
We employ groups of Large Language Models (LLMs) to solve innovation tasks, examining how they perform in isolation and how their social connectivity affects their collective behaviour. 


While many species exhibit culture~\citep{whiten1999CulturesChimpanzees,aplin2019CultureCulturalEvolution}, cultural change~\citep{aplin2015ExperimentallyInducedInnovations}, and even a continuously complexifying cultural repertoire~\citep{whiten2021EmergenceCollectiveKnowledge}, humans are unique in their ability to accumulate innovations~\citep{tennie2009RatchetingRatchetEvolution,derex2021HumanCumulativeCulture,boyd1996WhyCultureCommon}.
Studies aiming at understanding this phenomenon have drawn links between collective innovation and, among others, social learning capacities~\citep{dunbar_coevolution_1993,lotem_evolution_2017}, group size~\citep{dunbarCoevolutionNeocorticalSize1993,klinePopulationSizePredicts2010,derex2013ExperimentalEvidenceInfluence} and social connectivity~\citep{lazerNetworkStructureExploration2007,cantor2021SocialNetworkArchitecture,brackbillNetworkStructureCollective2017,fangBalancingExplorationExploitation2010,nisioti2022social}.

Despite a plurality of studies and methodologies, recent positions in cultural evolution (CE) warn that the field may have neglected certain mechanisms, such as the role of individual cognition in innovation and invention~\citep{singh2021SocialLearning,perry_not_2021,smolla_underappreciated_2021}.
The majority of collective innovation studies model individual inventions solely as random mutations and recombinations of existing innovations~\citep{masonPropagationInnovationsNetworked2008,cantor2021SocialNetworkArchitecture,fangBalancingExplorationExploitation2010}.
Yet a number of hypotheses in human studies point to the important role that language and, in general, advanced cognitive mechanisms, have played in our evolution~\citep{dunbarCoevolutionNeocorticalSize1993,boyd_evolution_2018,dunbar_how_2014,smith_cooperation_2017}.
Equipping models of CE with language can enable the study of such hypotheses and reduce the artificiality of experimental set-ups by bringing them closer to the ones used with human participants~\citep{mesoudi_experimental_2021}.

\begin{figure*}
\centering
        \includegraphics[width=0.95\textwidth]{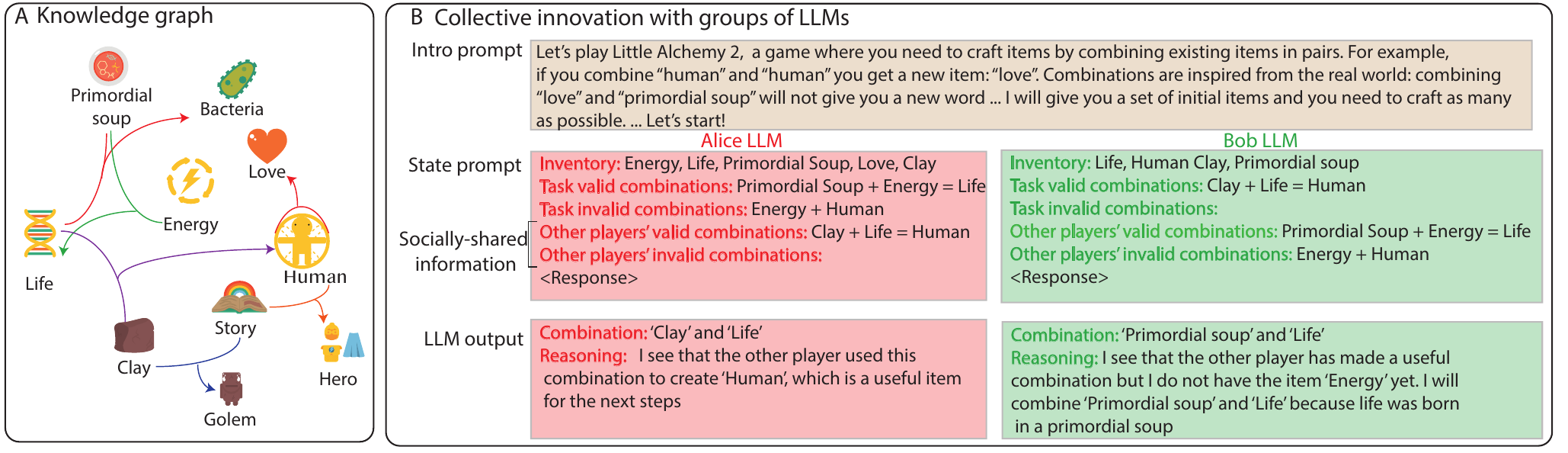}
    \caption{Studying collective innovation in groups of LLMs: A) we experiment with Little Alchemy 2 (\la), a game where players combine real-world items to create new ones. A knowledge graph describes the possible combinations (we only present a small sub-part of the graph which contains 720 items in total) B) Alice-LLM and Bob-LLM are two LLMs playing the game together.  
    They are provided with the same intro prompt, explaining the rules of the game, and the same task (they start with the same set of items). 
    Alice-LLM and Bob-LLM have identical weights but behave differently because the state prompt depends on their crafting history. 
    They are informed about the actions of others through their prompt. 
    In this paper, we study how groups of such LLM agents are able to efficiently explore a knowledge graph, focusing in particular on the effect of different social structures specifying with whom and when they can share information}
    \label{fig:intro}
\end{figure*}


Beyond their capacity to model human language, LLMs have shown an emergent ability to model certain aspects of human behaviour, such as fairness in economical decisions~\citep{horton_large_2023}, content biases in information transmission~\citep{acerbi_large_2023}, and convincing social interactions in realistic social simulation games~\citep{park_generative_2023}. Moreover, LLMs are evermore present as copilots of human labour and have become actors in the process human cultural evolution~\citep{brinkmann_machine_2023}.
A natural next question is whether they can also be useful in computational studies of cultural evolution as generative models of individuals~\citep{perez_cultural_2024}.

Here, we study how groups of LLMs solve collective innovation tasks.
As a test-bed, we employ Little Alchemy 2 (LA2)~\footnote{\url{https://littlealchemy2.com}}, an existing creative game where players need to combine items to create new ones.
To identify which combinations are valid, Little Alchemy 2 employs a knowledge graph 
where items are real-world entities and combinations are inspired by our physical reality (for example, 'fire' and 'water' results in a new item, 'steam'). Little Alchemy 2 was recently proposed as a test-bed for the study of human exploration, as it poses challenges not present in classically-employed bandit  tasks~\citep{brandle_empowerment_2023}.
Here, we test a similar proposal, namely that Little Alchemy 2 can be a useful test-best for studying both human and computational cultural evolution.
While Little Alchemy 2's knowledge graph is certainly not a comprehensive database of human cultural artefacts, it is significantly more realistic than previous test-beds containing a few hand-crafted interactions among symbolic items~\citep{derexPartialConnectivityIncreases2016,cantor2021SocialNetworkArchitecture,miglianoHuntergathererMultilevelSociality2020}. 
We present a visualization of a small part of the knowledge graph of \la in Figure \ref{fig:intro}.


First, we examine an LLM in isolation to probe its problem-solving capacities independently of group influences.
We identify three key challenges related to a) \textit{factual knowledge}: to efficiently explore the search space an LLM needs to leverage semantic knowledge about the task, b) \textit{multi-step reasoning}: to reach a target item a player often needs to craft multiple intermediate items, and c) \textit{exploration}: a key feature of \la is its open-ended nature. As is common in creative games, no instructions are given to the player, who starts with a couple of items and is left to explore a vast search space. As previous studies with human participants have shown, non-random exploration strategies, such as empowerment, are required to explore efficiently~\citep{brandle_empowerment_2023,klyubin_all_2005}. 
To our knowledge, the abilities of LLMs to exhibit such forms of exploration have not been studied before.



To examine the knowledge and multistep reasoning abilities of LLMs we first study innovation tasks that require crafting a target item.
These tasks were originally introduced by~\citet{jiangWordCraftEnvironmentBenchmarking2020}, and their complexity can be controlled by determining the number of intermediate items the player needs to craft before reaching the target, termed their depth.
Our experiments in tasks with a target suggest that: a) LLMs leverage factual/ knowledge, as removing the natural language semantics degrades performance b) multistep reasoning is challenging as performance significantly drops when increasing the depth of the task.

\begin{figure*}
\centering
\includegraphics[width=0.95\textwidth]{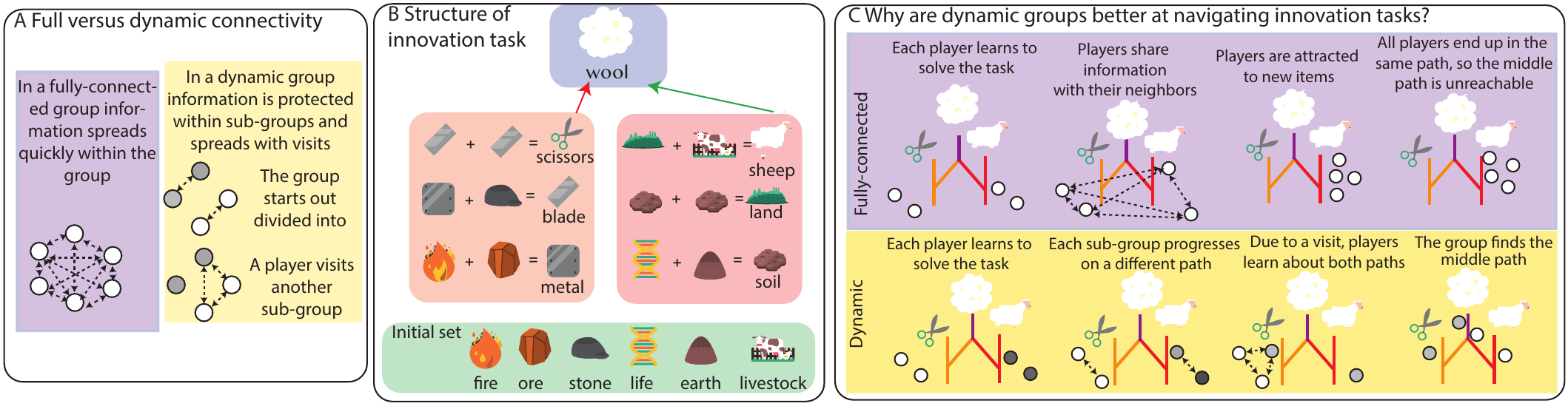}        \caption{Examining the effect of social connectivity on collective innovation: A) We consider two types of connectivity: a fully-connected group of 6 agents and a group with dynamic connectivity that starts out with agents divided into 3 sub-groups of two agents and visits of a fixed duration take place between groups with a random probability.
        B) Example structure of an innovation task in \la: the task starts out with 6 items that the player can combine to move up the search space.
        Depending on their semantics the items may create independent trajectories as the two presented here, culminating into the items "scissors" and "sheep".
        To discover the item "wool" the player needs to reach the end of both trajectories and combine them. 
    C) Why are dynamically-connected groups better at solving this innovation task? When a player observes that another player has found a new item it follows along. This means that, in a fully-connected group, all agents will get trapped in a single path. In contrast, subgroups of a dynamic group may explore different paths and, then, manage to recombine their solutions.
    } 
    \label{fig:deceptive}
\end{figure*}

We, then, focus on tasks without a target that employ the same graph and starting set of items with \la.
We, first, examine the ability of a single LLM to explore the search space and compare it with a baseline agent that employs empowerment and is known to perform on par with humans~\citep{brandle_empowerment_2023,klyubin_all_2005}.
Our experiments indicate that the LLM agent performs on par with an agent that randomly chooses combinations, suggesting that it does not explore efficienly.
Following the single-agent LLM experiments, we study the same tasks in a multi-LLM setting with 6 LLMs solving the same task. 
As we show on the right of Figure \ref{fig:intro}, LLMs in a group share information solely through the prompt: similarly to previous human lab studies~\citep{derexPartialConnectivityIncreases2016,miglianoHuntergathererMultilevelSociality2020}, the history of crafting actions of its neighbors is made available to the LLM. 
To study the effect of social structure, we compare the performance of a fully-connected group, where all agents communicate and share information with all other agents, and a group with dynamic connectivity employed in previous studies~\citep{derexPartialConnectivityIncreases2016,nisioti2022social,cantor2021SocialNetworkArchitecture} (we illustrate the two types of connectivity on the left of Figure \ref{fig:deceptive}).
In our multi-agent experiments, we observe that: a) LLMs learn imperfectly from social information, as there is some delay between the moment a neighbour of the LLM crafts an item and the LLM crafts it itself. We refer to this phenomenon as "imperfect copying" b) provided that they perfectly copy their neighbors, LLMs perform better in a collective than in isolation c) under the same assumption, groups with dynamic connectivity out-compete fully-connected groups.
This observation agrees with previous human and computational studies, which hypothesize that partially-connected groups are at an advantage due to the tree-like structure of innovation landscapes~\citep{derexPartialConnectivityIncreases2016,stanley_why_2015} (we illustrate this hypothesis on the right of Figure \ref{fig:deceptive}).





%% file: sections/related.tex
\section{Related work}
Various computational studies have investigated the dynamics of collective innovation. These studies differ across three important dimensions. First, they may model individual and social learning mechanisms differently.
The majority of CE models assumes that innovations arise solely through random mutations and recombinations~\citep{masonPropagationInnovationsNetworked2008,cantor2021SocialNetworkArchitecture,fangBalancingExplorationExploitation2010}, nullifying the cognitive capacities of individuals. Considering the problem-solving abilities of LLMs, discussed right after, our model can be seen as being rather closer to works that equip agents with cognition, for example through reinforcement learning~\citep{nisioti2022social}.
Second, studies propose different mechanisms as drivers of cultural accumulation.
Here, works may focus on the effect of social connectivity, with some suggesting that more connectivity is better and size is the sole determinant of collective performance~\citep{masonCollaborativeLearningNetworks2012}, while others suggesting that partial, static or dynamic, connectivity confers an advantage to groups exploring search spaces with local optima~\citep{derexPartialConnectivityIncreases2016,nisioti2022social,lazerNetworkStructureExploration2007}
Third, they employ different test-beds to capture the landscape of innovation. Initially computational studies studied classical search problems such as line search~\citep{masonPropagationInnovationsNetworked2008,masonCollaborativeLearningNetworks2012} and the NK-problem~\citep{lazerNetworkStructureExploration2007}, thus ignoring the hierarchical, tree-like structure of innovation landscapes. Another line of works employs a test-bed inspired from drug discovery, where individuals need to combine items to craft new ones~\citep{derexPartialConnectivityIncreases2016,cantor2021SocialNetworkArchitecture,nisioti2022social}, and is, thus, closest to the test-bed employed here. Yet this test-bed differs from ours in two ways: a) it is defined over symbolic items and manually designed small-scale knowledge graphs, thus failing to capture the scale and semantics of human innovation b) it assumes that each combination incurs a reward for the agent. In contrast, our open-ended tasks instruct the agent to explore and may, thus, capture the open-ended nature of human innovation~\citep{fogarty_cultural_2015}.


The cognitive capacities of LLMs have been under close scrutiny, with studies revealing surprising limitations considering their convincing conversational skills~\citep{mitchell_debate_2023}.
Such studies have focused on their factuality, planning and reasoning abilities, showing that LLMs are prone to hallucinations and falling into self-reinforcing loops~\citep{momennejad_evaluating_2023,wei_chain_2022}.
Multi-agent LLM studies~\citep{webb_prefrontal_2024,zhuge2023mindstorms} have revealed interesting phenomena reminiscent of collective intelligence with LLMs, such as that having them debate in a group prior to solving a task improves both their factuality and reasoning abilities~\citep{du_improving_2023,zhuge2023mindstorms}.
An understudied aspect of LLM cognition is, however, their ability to explore.
While LLMs have been employed for generating and evaluating goals for RL agents~\citep{colas2023augmenting,du_guiding_2023} and exploring in open-ended tasks~\citep{wang2023voyager}, such studies did not examine whether LLMs improved exploration due to leveraging semantic knowledge or planning.
When LLMs were evaluated on their ability to learn how to explore arbitrary bandits tasks solely through in-context learning, experiments indicated that only the state-of-the-art GPT-4 model, equipped with external memorization mechanisms, successfully engages in exploration~\citep{krishnamurthy_can_2024}.

%% file: sections/testbed.tex
\section{Collective innovation test-bed for LLMs}
We, here, describe the test-bed we have designed to study collective innovation with LLMs.
We have minimally extended Wordcraft, a Python-based gym reinforcement learing (RL) environment inspired by Little Alchemy 2 that was originally introduced for evaluating the commonsense abilities of RL agents~\citep{jiangWordCraftEnvironmentBenchmarking2020}.
Instead of images, items in Wordcraft are expressed as text.
Another difference with the original game is that tasks are targeted: whereas a game in Little Alchemy 2 starts with the player being given a small set of items and no further guidance, tasks in Wordcraft determine, in addition to the initial set, a target item to craft.
Our extension of Wordcraft is basically an introduction of open-ended tasks that follow the spirit of Little Alchemy 2 and an interface for mapping the gym environment to a textual form.
To capture the diversity of the task design space of our test-bed, below we
provide a general definition of its components and explain our particular implementation. These components are:

\vspace{-0.5cm}
\paragraph{A knowledge graph} that represents the semantics of the task space by indicating how items can be combined to create new items. In this work we employ the knowledge graph of Little Alchemy 2, but, in the general case, studies can manually define their own graphs~\citep{nisioti2022social} or derive graphs from text corpora~\citep{jiangWordCraftEnvironmentBenchmarking2020}.
\vspace{-0.5cm}

\paragraph{A task-generation process} that samples the initial set of items from the knowledge graph and determines the goal of the task. Here, we discriminate between open-ended and targeted tasks. The former prompt the LLM to discover as many items as possible without specifying a target. The latter prompt the LLM to craft a specific item and their complexity can be configured through two parameters: the number of items inserted in the initial set that are irrelevant for crafting the current target, $w$, termed \textit{distractors}, and the number of intermediate items that the LLM needs to craft before reaching the target, termed the \textit{depth}.
\vspace{-0.5cm}

\paragraph{A textual repesentation of the task}that has two parts: a) an intro prompt containing the rules of the game, as well as examples of tasks and correct outputs from the LLM in order to elicit in-context learning~\citep{brown_language_2020} b) the current task state. As a task requires multiple crafting steps (in the case of open-ended tasks on the scale of hundreds), we need a way to keep the prompt size limited. For this reason, instead of showing the complete crafting/discussion history for a task we summarize its current state in the following form for targeted tasks:

\begin{flushleft}
    \begin{tabular}{|p{0.95\columnwidth}|}
    \hline
    $<$Current task$>$
    
    \textit{Inventory}: set of items available for crafting
    
    \textit{Target}: target item
    
     \textit{Remaining rounds}: number of crafting steps before task is over
     
    \textit{Task valid combinations}: a list of combinations already attempted that gave a new item 
    
    \textit{Task invalid combinations}: a list of combinations already attempted that did not give a new item
 \\
    \hline
\end{tabular}
\end{flushleft}

In the case of open-ended tasks, information about the target is omitted. In addition to keeping the prompt size limited, this way of presenting the task can be seen as a form of external summarization~\citep{krishnamurthy_can_2024}, as the LLM does not need to memorize the item combinations it has attempted.
We provide an illustration of intro and task prompts in an example task where two LLMs are solving an open-ended task collectively on the right of Figure \ref{fig:intro}.

More formally, a task in our test-bed is described by a set of items that are initially available for crafting, $\mathcal{I}_0$, a knowledge graph $\mathcal{K}$ and a target item $g$ (that is empty in the case of open-ended tasks).
The initial set $\mathcal{I}$ and target $g$ are sampled from the knowledge graph $\mathcal{K}$ at the beginning of an episode that lasts for a fixed number of steps $T$.
The agent can execute actions in the form of two items, $[i_1, i_2]$ that denote the combination it attempts to make.
If, according to the knowledge graph, the action is a valid combination the environment returns a new item that is inserted in the inventory of the agent (if the item is already it is discarded, as items can be re-used indefinitely).
At each step $t$, the agent is presented with the current state of the environment and is prompted to execute an action.

%% file: sections/background.tex
\section{Baselines}
Here we describe previous models that do not make use of LLMs and  have been employed in single-agent and collective innovation studies.

\subsection{Singe-agent}
\paragraph{Empowered agent} 
Empowerment is a type of exploration strategy that centers around the idea of choosing actions that enable the generation of as many options as possible in the future~\citep{klyubin_all_2005}.
Previous lab studies where human participants played \la have shown that empowerment is a powerful exploration strategy in this game and that it captures human behavior more accurately than other exploration strategies, such as random exploration and uncertainty-minimization~\citep{brandle_empowerment_2023}.
While different implementations of empowerment are possible, here we consider the one employed in ~\citep{brandle_empowerment_2023}: 
at each timestep $t$ an agent employing empowerment combines the two items in its inventory that will result in crafting the most empowering item.
The empowerment of an item is computed as the number of valid combinations it participates in.
To choose an action, this agent accesses the knowledge graph of the task (defined in the previous section), computes the empowerment value of each potential combination and chooses the one with the highest value.
\vspace{-0.5cm}

\paragraph{Random agent} At each timestep $t$, this agent randomly combines two items from its inventory. 

\subsection{Multi-agent}
\paragraph{Random groups}
The majority of computational studies in collective innovation~\citep{masonPropagationInnovationsNetworked2008,cantor2021SocialNetworkArchitecture,fangBalancingExplorationExploitation2010, nisioti2022social} consider random agents (as defined in the previous paragraph) that can combine items they see in the inventories of their neighbours.
This implies that they have a perfect mechanism for copying social information.
Agents can introduce new items by randomly mutating a single item or combining multiple items.
For brevity, we refer to such groups of agents as random groups.
\vspace{-0.5cm}

\paragraph{Empowered groups}
Here we have multiple empowered agents that can combine items they see in the inventories of their neighbors.




%% file: sections/collective_LLM.tex
\section{Collective of LLMs with social connectivity}
Here, we describe how we designed group sof LLM agents for solving tasks in our innovation test-bed.
We first describe a single LLM agent and, then, describe how they share information in a group with a certain social connectivity.

At each timestep $t$ an LLM agent outputs text that contains the action it chooses to execute based on the prompt describing the task (described in the previous section). 
The prompt has been engineered to instruct the LLM to provide its output in the specific format:

\begin{flushleft}
    \begin{tabular}{|p{0.95\columnwidth}|}
    \hline
    \textit{Combination}: 'first item' and 'second item'
    
    \textit{Reasoning}: based on the information in the $<$Current task$>$, do reasoning about why you chose this combination
 \\
    \hline
\end{tabular}
\end{flushleft}

Instructing the LLM to follow a certain format is a common practice when it needs to interface with another system, such as an RL environment~\citep{wang_voyager_2023}.
Prompting the LLM to reason on its response is also a common technique, termed chain-of-thought prompting, and has been shown to improve the reasoning abilities of LLMs~\citep{wei_chain_2022}.
As the LLM output probabilities over tokens, we can control the randomness of its outputs through the temperature of a soft-max function over probabilities.

We consider groups of identical LLMs (they all have identical weights) that differ solely in the prompt that describes the task.
As we showed in Figure \ref{fig:intro}, the different LLMs are presented with the same intro prompt and are also assigned with the same task.
What differentiates LLMs is the state prompt, i.e., the current state of their inventory and their history of past combinations.
Thus, any differences in the outputs of LLMs in a given group solely arise due to their own behavior in the current task and are bootstrapped by the randomness in their sampling strategies.

A group is characterized by its social connectivity, an undirected graph $G$ that determines the local neighbourhood of an agent.
In this work, we consider two types of connectivity: a) in fully-connected groups all LLM agents are neighbours with each other b) in dynamic groups the group is divided into sub-groups of two and agents visit another random sub-group with probability $p$ for a fixed duration $V$ (see left of Figure \ref{fig:deceptive} for an illustration of the two connectivities).

Agents in a group interact solely by sharing information regarding their actions.
In particular, social information is shared by augmenting the prompt with the following tags:

\begin{flushleft}
    \begin{tabular}{|p{0.95\columnwidth}|}
    \hline
    \textit{Other players' valid combinations}: a list of combinations already attempted by other players in the agent's neighborhood that gave a new item 
    
    \textit{Other players' invalid combinations}: a list of combinations already attempted by other players in the agent's neighborhood that did not give a new item 
 \\
    \hline
\end{tabular}
\end{flushleft}

Thus, similarly to previous computational studies~\citep{masonPropagationInnovationsNetworked2008,cantor2021SocialNetworkArchitecture,fangBalancingExplorationExploitation2010, nisioti2022social}, information is shared implicitly based on the social connectivity without requiring an action on behalf of the agents.
We illustrate how socially-shared information is presented to the agents on the right of Figure \ref{fig:intro} and the two types of social connectivity that we consider in our study on the left of Figure \ref{fig:deceptive}.



%% file: sections/results.tex
\section{Results}

\subsection{Set-up}
We perform an experimental analysis where we control for different features of our set-up:
a) we consider both targeted and open-ended tasks. For the former we randomly sample tasks with different number of distractors ($w \in \{3,6\}$) and different depth values ($d \in \{1,2\}$).
We allow six crafting steps and measure success as the percentage of tasks where the agent crafts the target item.
We perform 10 trials and, for each one, we randomly sample 50 tasks.
Tasks are identical across trial.
Thus, variance within a trial reveals the effect of task variability while variance across trials reveals variability in the agent's policy for the same task.
For the open-ended tasks we employ the same initial set with the one used in Little Alchemy 2 ('air', 'earth','fire','water'), allow 200 crafting steps and average across 10 trials.
Here, we employ the size of the agent's inventory as a proxy for success
b) we study two different LLMs, \chatgpt~\footnote{\url{https://github.com/meta-llama/llama}} and an open-source model, \llama~\footnote{We use the 13-billion parameter model from \url{https://chat.openai.com/auth/login}} and compare their performance to the two single-agent baselines (random agents and agents using empowerment with a temperature of 0.1).
We employ a temperature value of 1.0 for the LLMs (we performed a grid search for \llama and did not observe large variations but expect that a high temperature value is useful for eliciting randomness across agents)
c) we study groups with two types of social connectivity: a fully-connected group with 6 agents and a dynamic group where agents are divided into subgroups of two and, unless otherwise specified, perform visits with a probability of $p=0.2$ that last for $V=$50 steps.
To avoid repetitions, if an agent repeats an already-attempted combination, we re-prompt until it chooses a novel one or 6 crafting steps have passed.
We provide code for reproducing experiments, including prompts, in \href{https://github.com/eleninisioti/CollectiveLLM.git}{an online repo}.

\subsection{LLMs can exploit task knowledge}
\begin{figure}
    \centering
    \includegraphics[width=0.9\columnwidth]{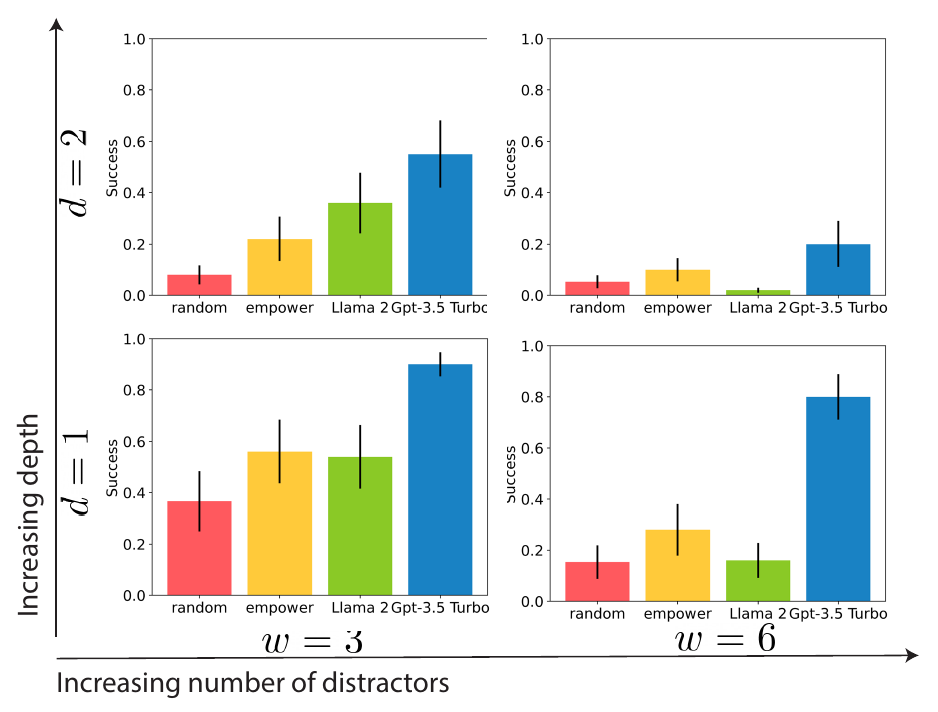}
    \caption{Performance in targeted tasks of varying complexity: success is the percentage of tasks for which the agent crafted the target item within the allowed time budget. }
    \label{fig:targeted}
\end{figure}

To gauge the ability of LLMs to understand and solve innovation tasks, we first examine the percentage of solved targeted tasks, $S$, for varying task complexity with single-agent methods in Figure~\ref{fig:targeted}.
Error bars indicate variance due to both task and agent variability.
We observe that when $w=3,d=1$ \chatgpt solves almost all tasks ($S=0.9 \pm 0.1$) while other methods succeed about half of the time. 
As these tasks contain 5 items, they can be solved successfully by exhaustive search within 10 steps and, by random search, about half of the time within the time budget.
Agents employing empowerment outperform random ones, as empowering items have higher chances of being targets.
\llama  performs significantly worse than \chatgpt, an observation that generalizes to all settings we examined.
A major reason for this is the inability of \llama to avoid repeating combinations.
In particular, we counted the number of times an LLM repeats a combination and observed that it was close to zero for \chatgpt but \llama repeats itself an overage of 4 times per task and in many cases, never finds a novel combination despite the repetition mechanism.

\begin{figure}
    \centering
    \includegraphics[width=0.9\columnwidth]{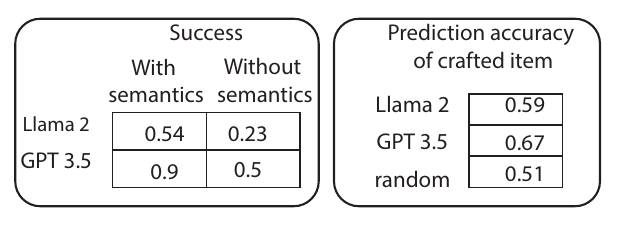}
    \caption{Examining the knowledge of LLMs in two probing tasks: (a) effect of removing semantics form the knowledge graph (b) semantic similarity (computed using glove embeddings) between the crafting outcomes predicted by the LLMs and the actual outcomes}
    \label{fig:knowledge}
\end{figure}

To understand these differences in performances, we perform two additional probing tasks that aim at examining the knowledge of LLMs.
First, we test to what extent the performance of LLMs is influenced by the semantics of the task.
We do so by encoding all words in the knowledge graph into random strings of 5 characters.
We present how success changes for the two LLMs when removing semantics in Table~\ref{fig:knowledge}, where we observe that the performance of both LLMs drops when semantics are removed.
The drop is much steeper for \chatgpt, which is not surprising, as it performs much better than \llama at the tasks with semantics.
The fact that semantics are crucial when there is one level indicates that the LLMs can predict which combination is the most likely to give the target item in a single step.

A yet more challenging and particularly useful skill, considering the multi-level nature of tasks, is predicting the outcome of crafting.
To test for this, we create another probing task: we prompt the LLM with valid combinations (generated by sampling combinations with the agent employing empowerment) and ask the LLM to predict the outcome of the combination.
We, then, compute the similarity between these predictions with the actual crafting outcome using the pre-trained glove model 'glove-twitter-25'~\footnote{\url{https://huggingface.co/Gensim/glove-twitter-25}}.
We present these results on the right of Figure~\ref{fig:knowledge}, where we include a baseline that randomly samples items from the knowledge graph as predictions for reference (similarity values range between 0 and 1).
We observe that \chatgpt is better than \llama at predicting crafting outcomes but is far from perfect (the \href{https://anonymous.4open.science/r/CollectiveLLM-7298/README.md}{online repo} includes details about this probing task, including the prompt and predictions of the LLMs).

\subsection{LLMs struggle at multi-step reasoning}
As we increase the complexity of the tasks we observe, in Figure \ref{fig:targeted} that:
a) when the number of distractors increases ($w=6,d=1$) the gap between \chatgpt and other methods increases. This is not surprising as \chatgpt exploits task knowledge best and the benefits of informed exploration become more apparent in larger search spaces
b) when the depth increases ($w=3,d=2$) then the performance drop is more significant and the disparity between the two LLM models decreases.
Combined with the observation that \chatgpt cannot perfectly predict outcomes this suggests that multi-step search poses a qualitatively different challenge that even advanced LLMs cannot solve.

\begin{figure}
    \centering
    \includegraphics[width=0.9\columnwidth]{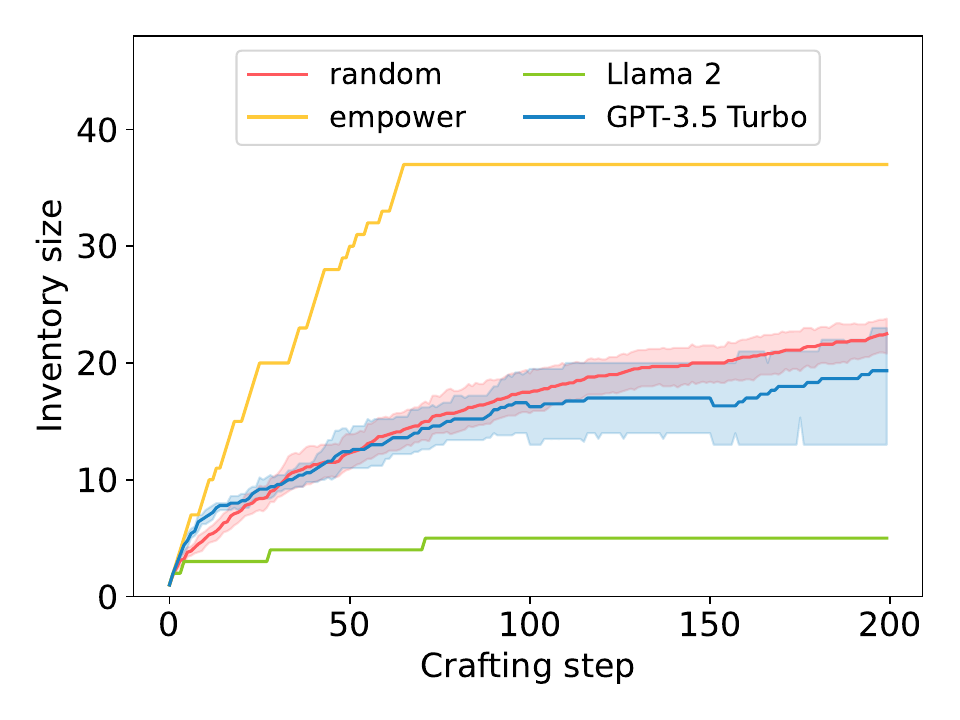}
    \caption{Performance in open-ended tasks for single agents}
    \label{fig:openended}
\end{figure}

\begin{figure}[t]
    \centering
    \includegraphics[width=0.9\columnwidth]{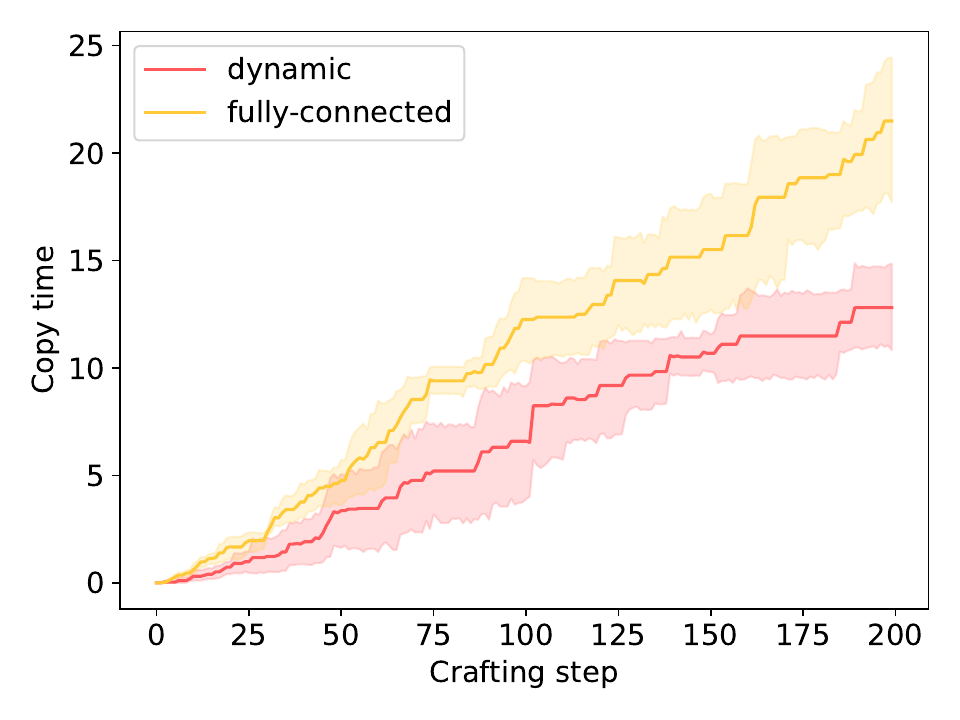}
    \caption{Copy time (number of timesteps it takes for a valid combination to appear in the inventory of an agent once it has appeared in one of its neighbors) for multi-agent \chatgpt}
    \label{fig:copy_time}
\end{figure}

\begin{figure*}
        \centering
    \includegraphics[width=0.9\textwidth]{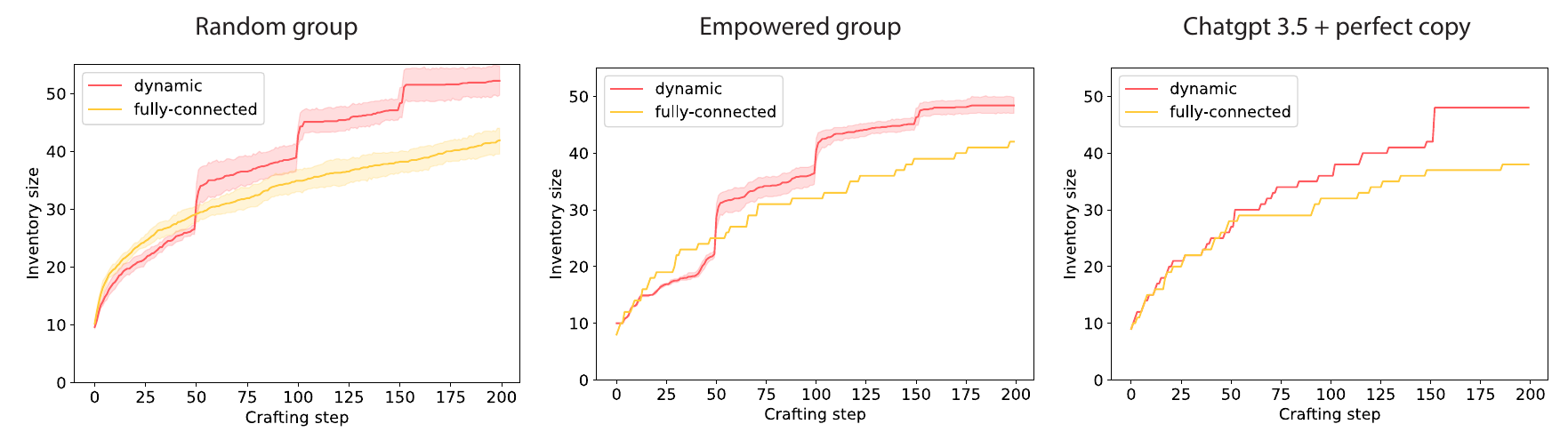}
    \caption{Effect of social connectivity in open-ended tasks with different agent models}
    \label{fig:connectivity}
\end{figure*}



\subsection{LLMs struggle in open-ended tasks}
We move to the open-ended task, where we observe that:
a) \llama only crafts an average of 6 items, performing worst among all methods. 
As discussed earlier, \llama has the tendency to repeat combinations  which is particularly detrimental in open-ended tasks due to the long horizon without resets
b) \chatgpt performs on par with random. 
This suggests that it does not leverage its knowledge for efficient exploration.
This model also repeats combinations but at a much smaller rate.
c) empowered agents perform best.
This is not surprising as empowered agents have access to the knowledge graph and
employ an exploration objective that is useful in such multi-level tasks~\citep{brandle_empowerment_2023}.

\subsection{LLMs copy others imperfectly}
We now move to experiments with groups where we first examine the collective behavior of \chatgpt agents (we do not experiment with \llama groups due to their sub-optimal performance).
A first question is whether LLM-agents benefit from innovating in a collective and, a requirement for this, is learning from the combinations of others (we have described how social information is shared when presenting the methods).
A proxy for this is the tendency of agents to copy actions they see in others.
To search for this we count the number of crafting steps it takes for an element to appear in the inventory of an agent once it appears in the inventory of one of its neighbors.
We compare the evolution of this variable for dynamic and fully-connected groups in Figure \ref{fig:copy_time} where we average within groups.
We observe that copying is not perfect: as time passes more and more items accumulate that the agent could have crafted.
Agents in fully-connected groups take more time to copy their neighbors.
Potential reasons for this are that copying when having more neighbors takes more time, as more items need to be copied, and that larger neighborhoods lead to longer prompts, making the task more difficult for LLMs.

\subsection{Connectivity influences collective innovation}
Finally, we compare groups with different connectivity in terms of their performance in open-ended tasks.
Figure \ref{fig:connectivity} reveals that, for all groups, dynamic connectivity performs best.
In the case of \chatgpt we present results with a perfect copy mechanism (when an agents crafts a new item, it appears in the inventories of all its neighbors).
As we saw in the previous paragraph, \chatgpt agents do not copy perfectly and are therefore less influenced by their connectivity (we performed the experiment with \chatgpt agents without a copy mechanism and did not see a difference between dynamic and fully-connected groups). 
Our empirical observation agrees with previous works that employed groups similar to our random group baseline ~\citep{masonPropagationInnovationsNetworked2008,cantor2021SocialNetworkArchitecture,fangBalancingExplorationExploitation2010}, humans~\citep{derexPartialConnectivityIncreases2016,miglianoHuntergathererMultilevelSociality2020} and reinforcement learning agents~\citep{nisioti2022social} and  were performed on manually-designed, small knowledge graphs.
By generalizing them to a larger knowledge graph grounded in the real world, our work further confirms that social structure matters in innovation tasks and suggests that our test-bed can prove useful in future studies of collective innovation with both human and artificial agents.

%% file: sections/discussion.tex
\section{Discussion}
We studied the ability of LLMs in innovation tasks, both in isolation and in groups with different social connectivity.
We have shown that \chatgpt can leverage the semantics of items to infer the outcomes of crafting but struggle when it comes to planning for multiple time steps and exploring in an open-ended way.
Nevertheless, groups of LLMs exhibit an interesting phenomenon previously found in human and computational studies: they perform better collectively when their social connectivity is partial and dynamic rather than static and fully-connected.
We attributed this phenomenon to the tree-like structure of the \la game: following down some paths may lead you away from other paths and slow down exploration.
We have shown that dynamically-connected LLM groups outperform both single LLMs, and fully-connected groups.
In groups with dynamic connectivity, subgroups explore different paths and exchange members that share information about other paths, increasing the diversity or breadth of exploration.

Our analysis focused on probing for specific abilities and emergent behaviours in the studied groups of LLMs. However, a larger-scale analysis is necessary to reveal the effect of the different hyperparameters of the model, such as the sampling strategy of LLMs, the configuration of the dynamic connectivity, and the task complexity.
A limitation revealed by our empirical study is that smaller, open-source models may fail at learning the task sufficiently well to lead to any interesting emergent behaviours.
Thus, as other studies of the cognitive capacities of LLMs have shown, our work suggests that collective innovation studies may require larger models, such as GPT-4 or the introduction of additional mechanisms for complementing their skills. 
We should note that our experiments did not examine whether pre-training equipped the LLMs with the ability to explore in-context, leverage common-sense knowledge or memorize the solution of Little Alchemy 2.
Nevertheless, our conclusion that groups with dynamic connectivity out-compete single-agent and fully-connected ones remains valid.

We believe that this work has important implications. We have shown that groups of LLMs with dynamic connectivity can overcome shortcomings of a single LLM and fully-connected groups. This is a key insight, as dynamic communication structures are less costly than fully-connected ones.
LLMs are becoming ubiquitous, participating in various human activities from finance to molecular discovery and writing fiction as copilots of human creativity and productivity~\citep{mirowski2022CoWritingScreenplaysTheatre,brinkmann_machine_2023}.
We believe that this work is a first step towards understanding how and with what kind of connectivity multi-agent LLM systems could optimally and efficiently participate in exploration, innovation, and cultural evolution. 
